\documentclass{article}

\usepackage{microtype}
\usepackage{graphicx}
\usepackage{subfigure}
\usepackage{booktabs} 

\usepackage{hyperref}

\usepackage[accepted]{icml2020}

\icmltitlerunning{Content-based Music Similarity with Triplet Networks}

\begin{document}

\twocolumn[
\icmltitle{Content-based Music Similarity with Triplet Networks}



\icmlsetsymbol{equal}{*}

\begin{icmlauthorlist}
\icmlauthor{Joseph Cleveland}{ic}
\icmlauthor{Derek Cheng}{cu}
\icmlauthor{Michael Zhou}{cu}
\icmlauthor{Thorsten Joachims}{cu}
\icmlauthor{Douglas Turnbull}{ic}
\end{icmlauthorlist}

\icmlaffiliation{ic}{Computer Science Department, Ithaca College, Ithaca, NY, USA}
\icmlaffiliation{cu}{Department of Computer Science, Cornell University, Ithaca, NY, USA}

\icmlcorrespondingauthor{Joseph Cleveland}{jcleveland@ithaca.edu}
\icmlcorrespondingauthor{Douglas Turnbull}{dturnbull@ithaca.edu}

\icmlkeywords{Machine Learning, ICML}

\vskip 0.3in
]



\printAffiliationsAndNotice{\icmlEqualContribution} 

\begin{abstract}
We explore the feasibility of using triplet neural networks to embed songs based on content-based music similarity. Our network is trained using triplets of songs such that two songs by the same artist are embedded closer to one another than to a third song by a different artist. We compare two models that are trained using different ways of picking this third song: at random vs. based on shared genre labels. Our experiments are conducted using songs from the Free Music Archive and use standard audio features. The initial results show that shallow Siamese networks can be used to embed music for a simple artist retrieval task.  
\end{abstract}

\vspace{-7mm}
\section{Introduction}



With the advent of commercial music streaming services like Spotify, Apple Music, and Soundcloud, it has become easier than ever for smaller, up and coming artists to publish and share their music. In general, these services make use of listening histories and user preferences when generating music recommendations for users \cite{barrington2009smarter}. However, these collaborative filtering-based (CF)  systems suffer the \emph{new-item cold-start} problem \cite{schedl2018current}: little or no historical user preference data exists for new or obscure artists and songs. Content-based (CB) approaches \cite{van2013deep}, on the other hand, offer an alternative that does not suffer from the cold-start problem.

In a related user study, we found that acoustic similarity plays an important part of how individuals construct playlists and recommend music to friends \cite{cheng2020}. As a result, we are interested in developing a model that can predict acoustic similarity for the purpose of music recommendation especially in cases where little or no preference information is available (i.e., long-tail music recommendation \cite{celma2008hits}). This work is motivated by our development of Localify\footnote[1]{https://www.localify.org}, a web-app for generating personalized playlists with local music.

In this paper, we explore the feasibility of using Triplet networks, a variant of Siamese networks \cite{bromley1994signature}, for content-based music recommendation. In this context, a Triplet network learns an embedding of an item such that the item is close to other similar items and far from dissimilar items in the embedding space. To train the network, we will consider songs by the same artist to be similar and songs by all other artists to be dissimilar.  We can then evaluate the quality of the embedding by how close songs of one artist are to each other when compared to songs by other artists. 

Triplet networks have received a good amount of recent attention for the task of facial recognition \cite{taigman2014deepface, schroff2015facenet} and have been applied in an number of different application domains (e.g., job-resume matching \cite{maheshwary2018matching}, speaker identification \cite{zeghidour2016joint}). 
In the context of music, \cite{park2017representation} have explored using Triplet networks for feature learning, genre classification, and content-based audio similarity. They used a deep convolutional architecture to create a useful embedding. In this work, we use a shallow fully connected network on a suite of audio features, seeking to create a system which would easily scale to a music recommendation scenario. We also explore the use genre information during model training. 

\vspace{-2mm}
\section{Triplet Networks}

Triplet networks output embeddings which are used to determine if two items are of the same class. They enable 1-shot learning where data of unseen classes can be compared. In our case, the audio content of two songs are compared to determine their similarity. Data are grouped into triplets containing an anchor song $a$, a positive song $p$, and a negative song $n$. We train the model such that the $l_2$ distance from the positive embedding $f(p)$ to the anchor embedding $f(a)$ is small, whereas $f(n)$ is far from $f(a)$. This is encouraged with the \emph{triplet loss function} \cite{schroff2015facenet}: 
\begin{eqnarray*}
\lefteqn{\mathcal{L}(a, p, n) =  }\\
 ~& \sum _{i=1}^{N} \left[ || f(a_i) - f(p_i) ||_2 ^ 2 - 
|| f(a_i) - f(n_i) ||_2 ^ 2 
+ \alpha \right]_+
\end{eqnarray*}
where $\alpha$ is the expected margin between positive and negative songs.  This function is minimized when the distance between $a$  and $p$ is less than the distance between $a$ and $n$. The network architecture in our case is a two-layer fully-connected network where the first layer uses a sigmoid activation, and the second layer uses a $tanh$ activation. All layers including the output are the same size as the input.

\section{Experiment Setup }

The experiments use the Free Music Archive (FMA) \cite{defferrard2016fma} dataset, specifically the 518 audio features from \path{features.csv} which they have extracted using Librosa \cite{mcfee2015librosa}. These features consist of every Librosa spectral feature, some of which are Zero-Crossing Rate, CQT, Tonnetz, MFCC, and STFT.

Triplets were generated where $a$ and $p$ are songs of the same artists, whereas $n$ is of a different artist. Distinguishing between songs from artists of different genres is often easy, so to train on more difficult examples, we trained a second Triplet network using a set of triplets that were generated where $a$ and $n$ are sampled from the same musical genre.

The Siamese networks were trained for 200 epochs of 512 randomly sampled triplets (102,400 unique triplets). The hyperparameter $\alpha$ for the margin was set to 1. Artists in the data set with only one song were filtered out, as it is impossible to match a unique anchor song with a unique positive song. Of the remaining 8,429 artists, all the songs from 70\% of these artists were sampled for training, and the songs from the remaining 30\% of these artists were used for evaluation. 

For our first baseline model, we evaluated the Euclidean ($l_2$) distance of the feature vectors in the raw feature space. Our second baseline is applying a z-score transformation to each of the 518 feature dimensions so that each dimension will have an expected mean of 0 and standard deviation of 1. We then calcuate $l_2$ distance in the z-scored transformed space.  

The Area Under the Receiver Operating Characteristic (AUC) was used as our evaluation metric \cite{manning2008introduction}. A perfect ranking with all relevant items at the top results in an AUC of 1.0 while randomly shuffling the ranking results in an expected AUC of 0.5. 

For each of the 2,529 artists in the evaluation set, an AUC score is calculated for one randomly selected \emph{query} song. The other songs by the artist of the query song are considered relevant while all other songs are not relevant.  The Euclidean ($l_2$) distance between the query song and all other songs in the evaluation set were calculated and used to rank them. We report the average AUC score for all artists in Table \ref{table:results}.

\section{Results }

Both Triplet networks outperform the baseline systems suggesting that the networks are learning an embedding that is helpful for the task of artist retrieval. Surprisingly, the Triplet network trained on randomly selected triplets performed best, followed by the network trained on triplets of the same genre. We had expected the opposite, as the triplets of the same genre are more difficult to distinguish which would hopefully make the model more robust. However, the random triplet model may have an advantage on this evaluation set.

\begin{table}[] \label{table:results}
\centering
\caption{The average AUC over 2,529 rankings of 25,290 songs for four embedding models: raw vectors and z-scored vectors baselines, and Triplet networks trained with two types of triplets.  }
\begin{tabular}{|c||c|}
\hline
 Model & AUC \\
 \hline\hline
 $l_2$ distance with raw vectors &  0.800\\
 $l_2$ distance with z-scored vectors  &  0.825\\
 Triplet NN (Random Triplets)  & \textbf{0.900}\\
 Triplet NN (Genre Triplets) &  0.883\\
 \hline
\end{tabular}
\vspace{-6mm}
\label{table:results}
\end{table}

\section{Discussion} 

\cite{park2017representation} showed that deep convolutional Triplet networks are good at creating embeddings which are useful for assessing audio similarity. We showed that this is also true of shallow Triplet networks. While our learned embeddings outperform simple similarity metrics, those simple metrics perform surprisingly well. This shows that even simple methods of analysis are viable for assessing similarity, but there may be a limit to the improvement one can make from transforming these simple acoustic features. 

Our strategy of using artists and genre labels to produce training data has very little cost, but may not be sufficient for creating a robust embedding model for content-based music recommendation. Many artists are acoustically diverse, making their songs poor examples for assessing acoustic similarity. In the future we would like to use similarity data from a listening study to assess the model. In lieu of creating all training data from a listening study, alternative methods of selecting hard triplets, such as those discussed in \cite{schroff2015facenet}, could be applied. 

There are many changes to model architecture, loss function, and data that need further exploration. Different suites of input features such as those available from the Spotify API could be compared, as well as different sizes of the embedding space. However, the proper creation of training and evaluation data is the most critical component to study, as there is much subjectivity in deciding the similarity of songs, particularly in the context of personalized recommendation.

This project is supported by NSF grant IIS-1615706/1615679 and IIS-1901168/1901330.

\bibliography{example_paper}
\bibliographystyle{icml2020}

\end{document}